\documentclass{esannV2}
\usepackage[dvips]{graphicx}
\usepackage[latin1]{inputenc}
\usepackage{amssymb,amsmath,array}
\usepackage{cite}

\usepackage{listings}
\usepackage{ifxetex}
\usepackage{fancyhdr}
\usepackage{hyperref}
\usepackage{graphicx}
\usepackage{wrapfig}
\usepackage{amsmath}
\usepackage{amsfonts}
\usepackage{amssymb}
\usepackage{amsthm}
\usepackage{listings}
\usepackage{setspace}
\usepackage{wrapfig}
\usepackage{tablefootnote}
\usepackage{multirow}
\usepackage{booktabs}
\usepackage[section]{placeins}
\usepackage{color}
\usepackage{amsfonts}
\usepackage{ulem}
\usepackage{algorithm}
\usepackage{algorithmicx}
\usepackage{algpseudocode}
\usepackage{graphicx}
\usepackage{float}
\usepackage{dsfont}
\usepackage{cleveref}

\allowdisplaybreaks

\newcommand{\mat}[1]{\mathbf{#1}}
\newcommand{\set}[1]{\mathcal{#1}}
\newcommand{\pnorm}[1]{\lVert{#1}\rVert}

\DeclareMathOperator*{\loss}{{\ell}}
\DeclareMathOperator*{\dist}{{d}}
\DeclareMathOperator*{\mad}{{MAD}}
\DeclareMathOperator*{\median}{{median}}
\DeclareMathOperator*{\trace}{{trace}}

\DeclareMathOperator*{\diag}{diag}
\DeclareMathOperator*{\sign}{sign}

\newcommand{\N}{\ensuremath{\mathcal{N}}} 
\newcommand{\I}{\mat{\mathbb{I}}}

\newcommand{\RN}{\mathbb{R}}

\DeclareMathOperator*{\regularization}{{\theta}}
\DeclareMathOperator*{\prototype}{{p}}
\newcommand{\x}{\ensuremath{\vec{x}}}
\newcommand{\y}{\ensuremath{y}}
\newcommand{\setx}{\ensuremath{\set{X}}}

\newcommand{\xcf}{\ensuremath{\vec{x}'}}
\newcommand{\ycf}{\ensuremath{y'}}
\newcommand{\classifier}{\ensuremath{h}}
\newcommand{\regression}{\ensuremath{f}}
\newcommand{\protolabel}{\ensuremath{o}}
\DeclareMathOperator*{\distmat}{{\mat{\Omega}}}

\begin{document}
\title{On the computation of counterfactual explanations - A survey}

\author{Andr\'e Artelt\footnote{corresponding author: \href{mailto:aartelt@techfak.uni-bielefeld.de}{aartelt@techfak.uni-bielefeld.de}}\; and Barbara Hammer
%
\thanks{We gratefully acknowledge funding from the VW-Foundation for project \textit{IMPACT} funded in the frame of the funding line \textit{AI and its Implications for Future Society}.}
%
\vspace{.3cm}\\
%
CITEC - Cognitive Interaction Technology \\
Bielefeld University - Faculty of Technology \\
Inspiration 1, 33619 Bielefeld - Germany
}

\maketitle

\begin{abstract}
Due to the increasing use of machine learning in practice it becomes more and more important to be able to explain the prediction and behavior of machine learning models. An instance of explanations are counterfactual explanations which provide an intuitive and useful explanations of machine learning models.

In this survey we review model-specific methods for efficiently computing counterfactual explanations of many different machine learning models and propose methods for models that have not been considered in literature so far.
\end{abstract}

\section{Introduction}
Due to recent advances in machine learning (ML), ML methods are increasingly use in real world scenarios~\cite{deeprl, deepsound, deepnlp, deepcv}. Especially, ML technology is nowadays used in critical situations like predictive policing~\cite{predictivepolicing} and loan approval~\cite{creditriskml}. In order to increase trust and acceptance of these kind of technology, it is important to be able to explain the behaviour and prediction of these models~\cite{molnar2019} - in particular answer questions like ``Why did the model do that? And why not smth. else?''. This becomes even more important in view to legal regulations like the EU regulation on GDPR~\cite{gdpr}, that grants the user a right to an explanation.

A popular method for explaining models~\cite{molnar2019, surveyxai, explainingexplanations, explainableartificialintelligence} are counterfactual explanations (often just called counterfactuals)~\cite{counterfactualwachter}. A counterfactual explanation states changes to some features that lead to a different (specified) behaviour or prediction of the model. Thus, counterfactual explanation can be interpreted as a recommendation what to do in order to achieve a requested goal. This is why counterfactual explanations are that popular - they are intuitive and user-friendly~\cite{molnar2019, counterfactualwachter}.

Counterfactual explanations are an instance of model-agnostic methods. Therefore, counterfactuals are not tailored to a particular model but can be computed for all possible models (in theory). Other instances of model-agnostic methods are feature interaction methods~\cite{featureinteraction}, feature importance methods~\cite{featureimportance}, partial dependency plots~\cite{partialdependenceplots} and local methods that approximates the model locally by an explainable model (e.g. a decisiontree)~\cite{lime2016, decisiontreecounterfactual}. The nice thing about model-agnostic methods is that they (in theory) do not need access to model internals and/or training data - it is sufficient to have an interface where we can pass data points to
the model and observe the output/predictions of the model.

However, it turns out that efficiently computing high quality counterfactual explanations of black-box models can be very difficult~\cite{counterfactualslvq}. Therefore, it is beneficial to develop model-specific methods - that use model internals - for efficiently computing counterfactual explanations. Whenever we have access to model internals, we can use the model-specific method over the model-agnostic method for efficiently computing counterfactual explanations. In this work we focus on such model-specific methods.

In particular, our contributions are:
\begin{itemize}
\item We review model-specific methods for efficiently computing counterfactual explanations of different ML models.
\item We propose model-specific methods for efficiently computing counterfactual explanations of models that have not been considered in literature so far.
\end{itemize}
The remainder of this paper is structured as follows: First, we briefly review counterfactual explanations (section~\ref{sec:counterfactualexplanations}). Then, in section~\ref{sec:computationcounterfactuals} we review and propose model-specific methods for computing counterfactual explanations. Finally, section~\ref{sec:conclusion} summarizes this papers. All derivations and mathematical details can be found in the appendix (section~\ref{sec:appendix}).

\section{Counterfactual explanations}\label{sec:counterfactualexplanations}
Counterfactual explanations~\cite{counterfactualwachter} (often just called counterfactuals) are an instance of example-based explanations~\cite{casebasedreasoning}. Other instances of example-based explanations~\cite{molnar2019} are influential instances~\cite{influentialinstances} and prototypes \& criticisms~\cite{prototypescriticism}.

A counterfactual states a change to some features/dimensions of a given input such that the resulting data point (called counterfactual) has a different (specified) prediction than the original input. Using a counterfactual instance for explaining the prediction of the original input is considered to be fairly intuitive, human-friendly and useful because it tells people what to do in order to achieve a desired outcome~\cite{counterfactualwachter,molnar2019}.

A classical use case of counterfactual explanations is loan application~\cite{creditriskml,molnar2019}:
\textit{Imagine you applied for a credit at a bank. Unfortunately, the bank rejects your application. Now, you would like to know why. In particular, you would like to know what would have to be different so that your application would have been accepted. A possible explanation might be that you would have been accepted if you would earn 500\$ more per month and if you would not have a second credit card.}

Although counterfactuals constitute very intuitive explanation mechanisms, there do exist a couple of problems.

One problem is that there often exist more than one counterfactual - this is called \textit{Rashomon effect}~\cite{molnar2019}. If there are more than one possible explanation (counterfactual), it is not clear which one should be selected.

An alternative - but very similar in the spirit - to counterfactuals~\cite{counterfactualwachter} is the Growing Spheres method~\cite{growingsphere}. However, this method suffers from the curse of dimensionality because it has to draw samples from the input space, which can become difficult if the input space is high-dimensional.\\\\
According to~\cite{counterfactualwachter}, we formally define the finding of a counterfactual as follows: Assume a prediction function $\classifier: \set{X} \mapsto \set{Y}$ is given. Computing a counterfactual $\xcf \in \RN^d$ of a given input $\x \in \RN^d$\footnote{We restrict ourself to $\RN^d$, but in theory one could use an arbitrary domain $\setx$.} can be interpreted as an optimization problem:
\begin{equation}\label{eq:counterfactualoptproblem}
\underset{\xcf \,\in\, \RN^d}{\arg\min}\; \loss\big(\classifier(\xcf), \ycf\big) + C \cdot \regularization(\xcf, \x)
\end{equation}
where $\loss()$ denotes a loss function that penalizes deviation of the prediction $\classifier(\xcf)$ from the requested prediction $\ycf$. $\regularization()$ denotes a regularization that penalizes deviations from the original input $\x$ and the hyperparameter $C$ denotes the regularization strength.

Two common regularizations are the weighted Manhattan distance and the generalized L2 distance. The weighted Manhattan distance is defined as
\begin{equation}\label{eq:weighted_l1}
\regularization(\xcf, \x) = \sum_j \alpha_j \cdot |(\x)_j - (\xcf)_j|
\end{equation}
where $\alpha_j > 0$ denote the feature wise weights. A popular choice~\cite{counterfactualwachter} for $\alpha_j$ is the inverse median absolute deviation of the $j$-th feature median in the training data set $\set{D}$:
\begin{equation}
\begin{split}
& \alpha_j = \frac{1}{{\mad}_j} \\
& \text{where }\\
& {\mad}_j = \underset{\x \,\in\, \set{D}}{\median}\left(\left|(\x)_j - \underset{\x \,\in\, \set{D}}{\median}\big((\x)_j\big)\right|\right)
\end{split}
\end{equation}
The weights $\alpha_j$ compensate for the (potentially) different variability of the features. However, because we need access to the training data set $\set{D}$, this regularization is not a truly model-agnostic method - it is not usable if we only have access to a prediction interface of a black-box model.

Although counterfactual explanations are a model-agnostic method, the computation of a counterfactual becomes much more efficient when having access to the internals of the model. In this work we assume that we have access to all needed model internals as well as access to the training data set - we will only need the training data for computing the weights $\alpha_j$ in the weighted Manhattan distance Eq.~\ref{eq:weighted_l1}. We do not need access to the training data if we do not use the weighted Manhattan distance or if we use some other methods for computing the weights $\alpha_j$ (e.g. setting all weights to $1$).\\\\
A slightly modified version of Eq.~\ref{eq:counterfactualoptproblem} was proposed in~\cite{counterfactualguidedbyprototypes}. The authors claim that the original formalization in Eq.~\ref{eq:counterfactualoptproblem} does not take into account that the counterfactual should lie on the data manifold - the counterfactual should be a plausible data instance. To deal with this issue, the authors propose to add two additional terms to the original objective Eq.~\ref{eq:counterfactualoptproblem}:
\begin{enumerate}
\item The distance/norm between the counterfactual $\xcf$ and the reconstructed version of it that has been computed by using a pretrained autoencoder.
\item The distance/norm between the encoding of the counterfactual $\xcf$ and the mean encoding of training samples that belong to the requested class $\ycf$.
\end{enumerate}
The first term is supposed to make sure that the counterfactual $\xcf$ lies on the data manifold and thus is a plausible data instance. The second term is supposed to accelerate the solver for computing the solution of the final optimization problem. Both claims have been evaluated empirically~\cite{counterfactualguidedbyprototypes}.

Recently, another approach for computing plausible/feasible counterfactual explanations was proposed~\cite{face}. Instead of computing a single counterfactual, the authors propose to compute a path of intermediate counterfactuals that lead to the final counterfactual. The idea behind this path of intermediate counterfactuals is to provide the user with a set of intermediate goals that finally lead to the desired goal - it might be more feasible to ``go into the direction" of the final goal step by step instead of accomplishing it in a single step. In order to compute such a path of intermediate counterfactuals, the authors propose different strategies for constructing a graph on the training data set - including the query point. In this graph, two samples are connected by a weighted edge if they are ``sufficient close to each other" - the authors propose different measurements for closeness (e.g. based on density estimation). The path of intermediate counterfactuals is equal to the shortest path between the query point and a point that satisfies the desired goal - this is the final counterfactual. Therefore the final counterfactual as well as all intermediate counterfactuals are elements from the training data set.

Despite the highlighted issues~\cite{counterfactualguidedbyprototypes, face} of the original formalization Eq.~\ref{eq:counterfactualoptproblem}, we stick to it and leave further investigations on the computation of feasible \& plausible counterfactuals as future research. However, many of the approaches for computing counterfactuals - that are discussed in this paper - can be augmented to restrict the space of potential counterfactuals. These restrictions provide an opportunity for encoding domain knowledge that lead to more plausible and feasible counterfactuals.

\section{Computation of counterfactuals}~\label{sec:computationcounterfactuals}
In the subsequent sections we explore model-specific methods for efficiently computing counterfactual explanations of many different ML models. But before looking at model-specific methods, we first (section~\ref{sec:generalcase}) discuss methods for dealing with arbitrary types of models - gradient based as well as gradient free methods.

Note that for the purpose of better readability and due to space constraints, we put all derivations in the appendix (section~\ref{sec:appendix}).

\subsection{The general case}\label{sec:generalcase}
We can compute a counterfactual explanation of any model we like by plugging the prediction function $\classifier$ of the model into Eq.~\ref{eq:counterfactualoptproblem} and choosing a loss (eq. 0-1 loss) and regularization (e.g. Manhattan distance) function. Depending on the model, loss and regularization function, the resulting optimization problem might be differentiable or not. If it is differentiable, we can use gradient-based methods like (L-)BFGS and conjugate gradients for solving the optimization problem. If Eq.~\ref{eq:counterfactualoptproblem} is not differentiable, we can use gradient-free methods like the Downhill-Simplex method or an evolutionary algorithm like CMA-ES or CERTIFAI~\cite{counterfactualsgeneticalgorithm} - the nice thing about evolutionary algorithms is that they can easily deal with categorical features. Another approach, limited to linear classifiers, for handling contious and discrete features is to use mixed-integer programming (MIP)~\cite{mipcounterfactual}. Unfortuantely, solving a MIP is NP-hard. However, there exist solvers that can compute an approximate solution very efficiently. Popular methods are branch-and-bound and branch-and-cut algorithms~\cite{mipsolver}. 

When developing model-specific methods for computing counterfactuals, we always consider untransformed inputs only - since a non-linear feature transformation usually makes the problem non-convex. Furthermore, we only consider the Euclidean distance and the weighted Manhattan distance as candidates for the regularization function $\regularization(\cdot)$.

\subsection{Separating hyperplane models}\label{sec:separatinghyperplane}
A model whose prediction function $\classifier$ can be written as:
\begin{equation}
\classifier(\x) = \sign(\vec{w}^\top\x + b)
\end{equation}
is called a separating hyperplane model. Popular instances of separating hyperplane models are SVM, LDA, perceptron and logistic regression.

Without loss of generality, we assume $\set{Y}=\{-1, 1\}$. Then, the optimization problem for computing a counterfactual explanation Eq.~\ref{eq:counterfactualoptproblem} can be rewritten as:
\begin{equation}\label{eq:cf:sephyperplane}
\begin{split}
& \underset{\xcf \,\in\,\RN^d}{\arg\min}\;\regularization(\xcf, \x) \\
& \text{s.t.}\\
& \vec{q}^\top\xcf + c < 0
\end{split}
\end{equation}
where
\begin{equation}
\vec{q} = -\ycf\vec{w}
\end{equation}
\begin{equation}
c = -b\ycf
\end{equation}
Depending on the regularization, the optimization problem Eq.~\ref{eq:cf:sephyperplane} becomes either a linear program (LP) - if the weighted Manhattan distance is used - or a convex quadratic program (QP) with linear constraints - if the Euclidean distance is used. More details can be found in the appendix (section~\ref{sec:appendix:sephyperplane}).

If we would have some discrete features instead of contious features only, we would obtain a MIP or MIQP as described in~\cite{mipcounterfactual}.

\subsection{Generalized linear model}\label{sec:glm}
In a generalized linear model we assume that the distribution of the response variable belongs to the exponential family. The expected value is connected to a linear combination of features by a link function, where different distributions have different link functions.

In the subsequent sections, we explore how to efficiently compute counterfactual explanations of popular instances of the generalized model.

\subsubsection{Logistic regression}\label{sec:logreg}
In logistic regression we model the response variable as a Bernoulli distribution. The prediction function $\classifier$ of a logistic regression model is given as
\begin{equation}
\classifier(\x) = \begin{cases} 1 & \quad \text{if } p(y=1 \mid \x) \geq t \\ -1  & \quad \text{otherwise} \end{cases}
\end{equation}
where $t$ is the discrimination threshold (often $t=0.5$) and
\begin{equation}
p(y=1 \mid \x) = \frac{1}{1 + \exp(-\vec{w}^\top\x - b)}
\end{equation}
When ignoring all probabilities and setting $t=0.5$, the prediction function $\classifier$ of a logistic regression model becomes a separating hyperplane:
\begin{equation}
\classifier(\x) = \sign(\vec{w}^\top\x + b)
\end{equation}
Therefore, computing a counterfactual of a logistic regression model is exactly the same as for a separating hyperplane model (section~\ref{sec:separatinghyperplane}).

\subsubsection{Softmax regression}\label{sec:softmaxref}
In softmax regression we model the distribution of the response variable as a generalized Bernoulli distribution. The prediction function $\classifier$ of a softmax regression model is given as:
\begin{equation}
\classifier(\x) = \underset{i \,\in\,\set{Y}}{\arg\max}\;\frac{\exp(\vec{w}_i^\top\x + b_i)}{\sum_k \exp(\vec{w}_k^\top\x + b_k)}
\end{equation}
In this case, the optimization problem for computing a counterfactual explanation Eq.~\ref{eq:counterfactualoptproblem} can be rewritten as:
\begin{equation}\label{eq:cf:softmaxreg}
\begin{split}
& \underset{\xcf \,\in\,\RN^d}{\arg\min}\;\regularization(\xcf, \x) \\
& \text{s.t.}\\
& \vec{q}_{ij}^\top\xcf + c_{ij} < 0 \quad \forall\,j \in \set{Y}, j\neq i=\ycf
\end{split}
\end{equation}
where
\begin{equation}
\vec{q}_{ij} = \vec{w}_j - \vec{w}_i
\end{equation}
\begin{equation}
c_{ij} = b_j - b_i
\end{equation}
Depending on the regularization, the optimization problem Eq.~\ref{eq:cf:softmaxreg} becomes either a LP - if the weighted Manhattan distance is used - or a convex QP with linear constraints - if the Euclidean distance is used. More information can be found in the appendix (section~\ref{sec:appendix:softmaxreg}).

\subsubsection{Linear regression}\label{sec:lr}
In linear regression we model the distribution of the response variable as a Gaussian distribution. The prediction function $\regression$ of a linear regression model is given as:
\begin{equation}
\regression(\x) = \vec{w}^\top\x + b
\end{equation}
The optimization problem for computing a counterfactual explanation Eq.~\ref{eq:counterfactualoptproblem} can be rewritten as:
\begin{equation}\label{eq:cf:lr}
\begin{split}
& \underset{\xcf \,\in\,\RN^d}{\arg\min}\;\regularization(\xcf, \x) \\
& \text{s.t.}\\
& \vec{w}^\top\xcf + c \leq \epsilon \\
& -\vec{w}^\top\xcf - c \leq \epsilon
\end{split}
\end{equation}
where
\begin{equation}
c = b - \ycf
\end{equation}
and $\epsilon \geq 0$ denotes the tolerated deviation from the requested prediction $\ycf$.

Depending on the regularization, the optimization problem Eq.~\ref{eq:cf:lr} becomes either a LP (if the weighted Manhattan distance is used) or a convex QP with linear constraints (if the Euclidean distance is used). More information can be found in the appendix (section~\ref{sec:appendix:lr}).

\subsubsection{Poisson regression}\label{sec:poissonreg}
In Poisson regression we model the distribution of the response variable as a Poisson distribution. The prediction function $\regression$ of a Poisson regression model is given as:
\begin{equation}
\regression(\x) = \exp(\vec{w}^\top\x + b)
\end{equation}
In this case, the optimization problem for computing a counterfactual explanation Eq.~\ref{eq:counterfactualoptproblem} can be rewritten as:
\begin{equation}\label{eq:cf:poissonreg}
\begin{split}
& \underset{\xcf \,\in\,\RN^d}{\arg\min}\;\regularization(\xcf, \x) \\
& \text{s.t.}\\
& \vec{w}^\top\xcf + c \leq \epsilon \\
& -\vec{w}^\top\xcf - c \leq \epsilon
\end{split}
\end{equation}
where
\begin{equation}
c = b - \log(\ycf)
\end{equation}
and $\epsilon \geq 0$ denotes the tolerated deviation from the requested prediction $\ycf$.

Depending on the regularization, the optimization problem Eq.~\ref{eq:cf:poissonreg} becomes either a LP (if the weighted Manhattan distance is used) or a convex QP with linear constraints (if the Euclidean distance is used). More information can be found in the appendix (section~\ref{sec:appendix:poissonreg}).

\subsubsection{Exponential regression}\label{sec:expreg}
In exponential regression we model the distribution of the response variable as a exponential distribution. The prediction function $\regression$ of an exponential regression model is given as:
\begin{equation}
\regression(\x) = -\frac{1}{\vec{w}^\top\x + b}
\end{equation}
Then, the optimization problem for computing a counterfactual explanation Eq.~\ref{eq:counterfactualoptproblem} can be rewritten as:
\begin{equation}\label{eq:cf:expreg}
\begin{split}
& \underset{\xcf \,\in\,\RN^d}{\arg\min}\;\regularization(\xcf, \x) \\
& \text{s.t.}\\
& \vec{w}^\top\xcf + c \leq \epsilon \\
& -\vec{w}^\top\xcf - c \leq \epsilon
\end{split}
\end{equation}
where
\begin{equation}
c = b + \frac{1}{\ycf}
\end{equation}
and $\epsilon \geq 0$ denotes the tolerated deviation from the requested prediction $\ycf$.

Depending on the regularization, the optimization problem Eq.~\ref{eq:cf:expreg} becomes either a LP (if the weighted Manhattan distance is used) or a convex QP with linear constraints (if the Euclidean distance is used). More information can be found in the appendix (section~\ref{sec:appendix:expreg}).

\subsection{Gaussian naive Bayes}\label{sec:gaussnb}
The Gaussian naive Bayes model makes the assumption that all features are independent of each other and follow a normal distribution. The prediction function $\classifier$ of a Gaussian naive Bayes model is given as:
\begin{equation}
\classifier(\x) = \underset{i \,\in\,\set{Y}}{\arg\max}\;\prod_{k=1}^d \N(\x \mid \mu_{ik}, \sigma^2_{ik})\pi_i
\end{equation}
where $\pi_i$ denotes the a-priori probability of the $i$-th class.

The optimization problem for computing a counterfactual explanation Eq.~\ref{eq:counterfactualoptproblem} can be rewritten as:
\begin{equation}\label{eq:cf:gnb}
\begin{split}
& \underset{\xcf \,\in\,\RN^d}{\arg\min}\;\regularization(\xcf, \x) \\
& \text{s.t.}\\
& \xcf^\top\mat{A}_{ij}\xcf + \vec{q}_{ij}^\top\xcf + c_{ij} < 0 \quad\forall\,j\in\set{Y},j\neq i=\ycf
\end{split}
\end{equation}
where
\begin{equation}
\mat{A}_{ij} = \diag\left(\frac{1}{2\sigma_{ik}^2} - \frac{1}{2\sigma_{jk}^2}\right)
\end{equation}
\begin{equation}
\vec{q}_{ij} = \left(\frac{\mu_{j1}}{\sigma_{j1}^2} - \frac{\mu_{i1}}{\sigma_{i1}^2}, \dots, \frac{\mu_{jd}}{\sigma_{jd}^2} - \frac{\mu_{id}}{\sigma_{id}^2}\right)^\top
\end{equation}
\begin{equation}
c_{ij} = \log\left(\frac{\pi_j}{\pi_i}\right) + \sum_{k=1}^d \log\left(\frac{\sqrt{2\pi\sigma_{ik}^2}}{\sqrt{2\pi\sigma_{jk}^2}}\right) - \frac{\mu_{jk}^2}{2\sigma_{jk}^2} + \frac{\mu_{ik}^2}{2\sigma_{ik}^2}
\end{equation}
Because we can not make any statement about the definiteness of $\mat{A}_{ij}$, the quadratic constraints in Eq.~\ref{eq:cf:gnb} are non-convex. Therefore, the optimization problem Eq.~\ref{eq:cf:gnb} is a non-convex quadratically constrained quadratic program (QCQP).

We can approximately solve Eq.~\ref{eq:cf:gnb} by using an approximation method like the Suggest-Improve framework~\cite{park2017general}. Furthermore, if we have a binary classification problem, we can solve a semi-definite program (SDP) whose solution is equivalent to Eq.~\ref{eq:cf:gnb}. More details can be found in the appendix (sections~\ref{sec:appendix:gnb},\ref{sec:appendix:nonconvexqcqp} and~\ref{sec:appendix:qcqp1constraint}).

\subsection{Quadratic discriminant analysis}\label{sec:qda}
In quadratic discriminant analysis (QDA) we model each class distribution as an independent Gaussian distribution - note that in contrast to LDA each class distribution has its own covariance matrix. The prediction function $\classifier$ of a QDA model is given as:
\begin{equation}
\classifier(\vec{x}) = \underset{i \,\in\,\set{Y}}{\arg\max}\;\N(\vec{x} \mid \vec{\mu}_i, \mat{\Sigma}_i)\pi_i
\end{equation}
where $\pi_i$ denotes the a-priori probability of the $i$-th class.

In this case, the optimization problem for computing a counterfactual explanation Eq.~\ref{eq:counterfactualoptproblem} can be rewritten as:
\begin{equation}\label{eq:cf:qda}
\begin{split}
& \underset{\xcf \,\in\,\RN^d}{\arg\min}\;\regularization(\xcf, \x) \\
& \text{s.t.}\\
& \frac{1}{2}\xcf^\top\mat{A}_{ij}\xcf + \xcf^\top\vec{q}_{ij} + c_{ij} < 0 \quad\forall\,j\in\set{Y},j\neq i=\ycf
\end{split}
\end{equation}
where
\begin{equation}
\mat{A}_{ij} = \mat{\Sigma}_i^{-1} - \mat{\Sigma}_j^{-1}
\end{equation}
\begin{equation}
\vec{q}_{ij} = \mat{\Sigma}_j^{-1}\vec{\mu}_j - \mat{\Sigma}_i^{-1}\vec{\mu}_i
\end{equation}
\begin{equation}
c_{ij} = \frac{1}{2}\Big(\vec{\mu}_i^\top\mat{\Sigma}_i^{-1}\vec{\mu}_i - \vec{\mu}_j^\top\mat{\Sigma}_j^{-1}\vec{\mu}_j\Big) + \frac{1}{2}\log\left(\frac{\det(\mat{\Sigma}_i)}{\det(\mat{\Sigma}_j)}\right) + \log\left(\frac{\pi_j}{\pi_i}\right)
\end{equation}
Because we can not make any statement about the definiteness of $\mat{A}_{ij}$, the quadratic constraints in Eq.~\ref{eq:cf:qda} are non-convex. Thus, like in  Gaussian naive Bayes (section~\ref{sec:gaussnb}), the optimization problem Eq.~\ref{eq:cf:qda} is a non-convex QCQP.

Like in the case of the previous non-convex QCQPs, we can approximately solve Eq.~\ref{eq:cf:qda} by using an approximation method. Furthermore, if we have a binary classification problem, we can solve a SDP whose solution is equivalent to Eq.~\ref{eq:cf:qda}. More details can be found in the appendix (sections~\ref{sec:appendix:qda},\ref{sec:appendix:nonconvexqcqp} and~\ref{sec:appendix:qcqp1constraint}).

\subsection{Learning vector quantization models}\label{sec:lvq}
Learning vector quantization (LVQ) models~\cite{lvqreview} compute a set of labeled prototypes $\{(\vec{\prototype}_i, \protolabel_i)\}$ from a given training data set - we refer to the $i$-th prototype as $\vec{\prototype}_i$ and the corresponding label as $\protolabel_i$. The prediction function $\classifier$ of a LVQ model is given as:
\begin{equation}\label{eq:lvq_predict}
\begin{split}
& \classifier(\x) = \protolabel_i \\
& \text{s.t. } \min\,\dist(\x, \vec{\prototype}_i)
\end{split}
\end{equation}
where $\dist()$ denotes a function for computing the distance between a data point and a prototype - usually this is the Euclidean distance:
\begin{equation}\label{eq:distfunction}
\dist(\x, \vec{\prototype}) = (\x - \vec{\prototype})^\top \I (\x - \vec{\prototype})
\end{equation}
There exist LVQ models like (L)GMLVQ~\cite{gmlvq} and (L)MRSLVQ~\cite{mrslvq} that learn a custom (class or prototype specific) distance matrix $\distmat_{\prototype}$ that is used instead of the identity $\I$ when computing the distance between a data point and a prototype. This gives rise to the generalized L2 distance:
\begin{equation}\label{eq:lvq_generaldist}
\dist(\x, \vec{\prototype}) = (\x - \vec{\prototype})^\top{\distmat}_{\prototype}(\x - \vec{\prototype})
\end{equation}
Because a LVQ model assigns the label of the nearest prototype to a given input, the nearest prototype of a counterfactual must be a prototype $\vec{\prototype}_i$ with $\protolabel_i=\ycf$. According to~\cite{counterfactualslvq}, for computing a counterfactual, it is sufficient to solve the following optimization problem for each prototype $\vec{\prototype}_i$ with $\protolabel_i=\ycf$ and select the counterfactual $\xcf$ yielding the smallest value of $\regularization(\xcf, \x)$:
\begin{equation}\label{eq:cflvq_general}
\begin{split}
& \underset{\xcf \,\in\, \RN^d}{\arg\min}\;\regularization(\xcf, \x) \\
& \text{s.t.} \\
& \dist(\xcf, \vec{\prototype}_i) < \dist(\xcf, \vec{\prototype}_j) \quad \forall\, \vec{\prototype}_j\in\set{P}(\ycf)
\end{split}
\end{equation}
where $\set{P}(\ycf)$ denotes the set of all prototypes \emph{not} labeled as $\ycf$. Note that the feasible region of Eq.~\ref{eq:cflvq_general} is always non-empty - the prototype $\vec{\prototype}_i$ is always a feasible solution.

In the subsequent sections we explore the type of constraints of Eq.~\ref{eq:cflvq_general} for different LVQ models.

\subsubsection{(Generalized matrix) LVQ}\label{sec:gmlvq}
In case of a (generalized matrix) LVQ model - all prototypes use the same distance matrix $\distmat$, the optimization problem Eq.~\ref{eq:cflvq_general} becomes~\cite{counterfactualslvq}:
\begin{equation}\label{eq:cf:gmlvq}
\begin{split}
& \underset{\xcf \,\in\,\RN^d}{\arg\min}\;\regularization(\xcf, \x)\\
& \text{s.t.} \\
& \xcf^\top\vec{q}_{ij} + c_{ij} < 0 \quad \forall\, \vec{\prototype}_j\in\set{P}(\ycf)
\end{split}
\end{equation}
where
\begin{equation}
\vec{q}_{ij} = \frac{1}{2}\distmat(\vec{\prototype}_{j} - \vec{\prototype}_i)
\end{equation}
\begin{equation}
c_{ij} =  \frac{1}{2}\left(\vec{\prototype}_{i}^\top\distmat\vec{\prototype}_{i} - \vec{\prototype}_{j}^\top\distmat\vec{\prototype}_{j}\right)
\end{equation}
Depending on the regularization, the optimization problem Eq.~\ref{eq:cf:gmlvq} becomes either a LP (if the Euclidean distance is used) or a convex QP with linear constraints (if the weighted Manhattan distance is used). More information can be found in the appendix (section~\ref{sec:appendix:lvq}).

\subsubsection{(Localized generalized matrix) LVQ}\label{sec:lgmlvq}
In case of a (localiced generalized matrix) LVQ model - there are different, class or prototype specific, distance matrices $\distmat_p$, the optimization problem Eq.~\ref{eq:cflvq_general} becomes~\cite{counterfactualslvq}:
\begin{equation}\label{eq:cf:lgmlvq}
\begin{split}
& \underset{\xcf \,\in\,\RN^d}{\arg\min}\;\regularization(\xcf, \x) \\
& \frac{1}{2}\xcf^\top\mat{A}_{ij}\xcf + \xcf^\top\vec{q}_{ij} + c_{ij} < 0 \quad \forall\, \vec{\prototype}_j\in\set{P}(\ycf)
\end{split}
\end{equation}
where
\begin{equation}
\mat{A}_{ij} = {\distmat}_{i} - {\distmat}_{j}
\end{equation}
\begin{equation}
\vec{q}_{ij} = \frac{1}{2}\left({\distmat}_j\vec{\prototype}_{j} - {\distmat}_i\vec{\prototype}_{i}\right)
\end{equation}
\begin{equation}
c_{ij} = \frac{1}{2}\left(\vec{\prototype}_{i}^\top{\distmat}_{i}\vec{\prototype}_{i} - \vec{\prototype}_{j}^\top{\distmat}_{j}\vec{\prototype}_{j}\right)
\end{equation}
Because we can not make any statement about the definiteness of $\mat{A}_{ij}$, the quadratic constraints in Eq.~\ref{eq:cf:lgmlvq} are non-convex. Thus, like in  Gaussian naive Bayes (section~\ref{sec:gaussnb}) and QDA (section~\ref{sec:qda}), the optimization problem Eq.~\ref{eq:cf:lgmlvq} is a non-convex QCQP.

Like the previous non-convex QCQPs, we can approximately solve Eq.~\ref{eq:cf:lgmlvq} by using an approximation method. Furthermore, if we have a binary classification problem and each class is represented by a single prototype, we can solve a SDP whose solution is equivalent to Eq.~\ref{eq:cf:lgmlvq}. More details can be found in the appendix (sections~\ref{sec:appendix:lvq},\ref{sec:appendix:nonconvexqcqp} and~\ref{sec:appendix:qcqp1constraint}).

\subsection{Tree based models}\label{sec:treemodel}
Tree based models are very popular in data science because they often achieve a high predictive-accuracy~\cite{treebasedmodels}. In the subsequent sections we discuss how to compute counterfactual explanations of tree based models. In particular, we consider decision/regression trees and tree based ensembles like random forest models.

\subsubsection{Decision trees}\label{sec:decisiontree}
In case of decision/regression tree models, we can compute a counterfactual by enumerating all possible paths that lead to the requested prediction~\cite{decisiontreecounterfactual, interprettreeesembles}. However, it might happen that some requested predictions are not possible because all possible predictions of the tree are encoded in the leafs. In this case one might define an interval of acceptable predictions so that a counterfactual exists.

The procedure for computing a counterfactual of a decision/regression tree is described in Algorithm~\ref{algo:cf:tree}.
\begin{algorithm}[!htb]
\caption{Computing a counterfactual of a decision/regression tree}\label{algo:cf:tree}
\textbf{Input:} Original input $\x$, requested prediction $\ycf$ of the counterfactual, the tree model\\
\textbf{Output:} Counterfactual $\xcf$
\begin{algorithmic}[1]
 \State Enumerate all leafs with prediction $\ycf$
 \State For each leaf, enumerate all paths reaching the leaf
 \State For each path, compute the minimal change to $\x$ that yields the path
 \State Sort all paths according to regularization of the change to $\x$
 \State Select the path and the corresponding change to $\x$ that minimizes the regularization
\end{algorithmic}
\end{algorithm}

\subsubsection{Tree based ensembles}\label{sec:randomforest}
Popular instances of tree based enesmbles are random forest and gradient boosting regression trees. It turns out that the problem of computing a counterfactual explanation of such models is NP-hard~\cite{interprettreeesembles}.

The following heuristic for computing a counterfactual explanation of a random forest model was proposed in~\cite{ceml}: First, we compute a counterfactual of a model from the ensemble. Next, we use this counterfactual as a starting point for minimizing the number of trees that do not outpur the requested prediction by using a gradient-free optimization method like the Downhill-Simplex method. The idea behind this approach is that the counterfactual of a tree from the ensemble is close to the decision boundary of the ensemble so that computing a counterfactual of the ensemble becomes easier. By doing this for all trees in the ensemble, we get many counterfactuals and we can select the one that minimizes the regularization the most. This heuristic seems to work well in practice~\cite{ceml}.

Another approach for computing counterfactual explanations of an ensemble of trees was propsed in~\cite{interprettreeesembles} - although the authors do not call it counterfactuals, they actually compute counterfactuals. Their algorithm works as follows:
We iterate over all trees in the ensemble that do not yield the requested prediction. Next, we compute all possible counterfactuals of each of these trees (see section~\ref{sec:decisiontree}). If this counterfactual turns our to be counterfactual of the ensemble, we store it so that in the end we can select the counterfactual with the smallest deviation from the original input. However, it can not be guaranteed that a counterfactual of the ensemble is found because it might happen that by changing the data point so that it becomes a counterfactual of a particular tree, the prediction of other trees in the ensemble change as well. According to the authors, this algorithm/heuristic works well in practice. Unfortunately, the worst-case complexity is exponential in the number of features and thus it is not suitable for high dimensional data.

\section{Implementation}
The gradient-based and gradient free methods, as well as the model specific methods for tree based models are already implemented in CEML~\cite{ceml}. The implementation of the LVQ specific methods are provided by the authors of~\cite{counterfactualslvq}. The Python~\cite{van1995python} implementation of our proposed methods is available on GitHub\footnote{\url{https://github.com/andreArtelt/OnTheComputationOfCounterfactualExplanations}} and is based on the Python packages scikit-learn~\cite{scikit-learn}, numpy~\cite{numpy} and cvxpy~\cite{cvxpy}.

We plan to add these model-specific methods to CEML~\cite{ceml} in the near future.

\section{Conclusion}\label{sec:conclusion}
In this survey we extensively studied how to compute counterfactual explanations of many different ML models. We reviewed known methods from literature and proposed methods (mostly LPs and (QC)QPs) for computing counterfactuals of ML models that have not been considered in literature so far.

\section{Appendix}\label{sec:appendix}
\subsection{Relaxing strict inequalities}\label{sec:appendix:relaxingstrictinequlities}
When modeling the problem of computing counterfactuals, we often obtain strict inequalities like
\begin{equation}
g(\x) < 0
\end{equation}
Strict inequalities are not allowed in convex programming because the feasible region would become an open set. However, we could turn the $<$ into a $\leq$ by adding a small number  to the left side of the inequality:
\begin{equation}
g(\x) + \epsilon \leq 0
\end{equation}
where $\epsilon > 0$ is a small number.

In practice, when implementing our methods, we found that we can often safely replace all $<$ by $\leq$ without changing anything else - this might be because of the numerics (like round-off errors) of fixed size floating-point numbers.

\subsection{Separating hyperplane}\label{sec:appendix:sephyperplane}
Recall that the prediction function $\classifier$ is given as:
\begin{equation}
\classifier(\x) = \sign(\vec{w}^\top\x + b)
\end{equation}
If we multiply the projection $\vec{w}^\top\x + b$ by the requested prediction $y$\footnote{Note that we assume $\set{Y}=\{-1, 1\}$.}, the result is positive if and only if the classification $\classifier(\x)$ is equal to $y$. Therefore, the linear constraint for predicting class $y$ is given as
\begin{equation}
\begin{split}
& y\left(\vec{w}^\top\x + b\right) > 0 \\
& \Leftrightarrow \vec{q}^\top\x + c < 0
\end{split}
\end{equation}
where
\begin{equation}
\vec{q} = -y\vec{w}
\end{equation}
\begin{equation}
c = -b\y
\end{equation}

\subsection{Generalized linear models}\label{sec:appendix:glm}
\subsubsection{Softmax regression}\label{sec:appendix:softmaxreg}
Recall that the prediction function $\classifier$ is given as:
\begin{equation}
\classifier(\x) = \underset{i \,\in\,\set{Y}}{\arg\max}\;\frac{\exp(\vec{w}_i^\top\x + b_i)}{\sum_k \exp(\vec{w}_k^\top\x + b_k)}
\end{equation}
Thus, the constraint for obtaining a specific prediction $\ycf$ is given as:
\begin{equation}\label{eq:softmaxreg:constraint}
\frac{\exp(\vec{w}_i^\top\x + b_i)}{\sum_k \exp(\vec{w}_k^\top\x + b_k)} > \frac{\exp(\vec{w}_j^\top\x + b_j)}{\sum_k \exp(\vec{w}_k^\top\x + b_k)} \quad \forall\,j\neq i=\ycf
\end{equation}
Holding $i$ and $j$ fixed, we can simplify Eq.~\ref{eq:softmaxreg:constraint}:
\begin{equation}
\begin{split}
\frac{\exp(\vec{w}_i^\top\x + b_i)}{\sum_k \exp(\vec{w}_k^\top\x + b_k)} &> \frac{\exp(\vec{w}_j^\top\x + b_j)}{\sum_k \exp(\vec{w}_k^\top\x + b_k)} \\
& \Leftrightarrow \\ \exp(\vec{w}_i^\top\x + b_i) &> \exp(\vec{w}_j^\top\x + b_j) \\
& \Leftrightarrow \\ \vec{w}_i^\top\x + b_i &> \vec{w}_j^\top\x + b_j \\
& \Leftrightarrow \\ \vec{q}_{ij}^\top\xcf + c_{ij} &< 0
\end{split}
\end{equation}
where
\begin{equation}
\vec{q}_{ij} = \vec{w}_j - \vec{w}_i
\end{equation}
\begin{equation}
c_{ij} = b_j - b_i
\end{equation}
Therefore, we can rewrite Eq.~\ref{eq:softmaxreg:constraint} as a set of linear inequalities.

\subsubsection{Linear regression}\label{sec:appendix:lr}
Recall that the prediction function $\regression$ is given as:
\begin{equation}
\regression(\x) = \vec{w}^\top\x + b
\end{equation}
By introducing the parameter $\epsilon \geq 0$ that specifies the maximum tolerated deviation from the requested prediction - we set $\epsilon = 0$ if we do not allow any deviations - the constraint for obtaining the requested prediction $\ycf$ is given as
\begin{equation}\label{eq:lr:constraint}
\begin{split}
& |\regression(\xcf) - \ycf| \leq \epsilon \\
& \Leftrightarrow |\vec{w}^\top\xcf + b - \ycf| \leq \epsilon \\
& \Leftrightarrow |\vec{w}^\top\xcf + c| \leq \epsilon 
\end{split}
\end{equation}
where
\begin{equation}
c = b - \ycf
\end{equation}
Finally, we can rewrite Eq.~\ref{eq:lr:constraint} as two linear inequality constraints:
\begin{equation}
\begin{split}
& \vec{w}^\top\xcf + c \leq \epsilon \\
& -\vec{w}^\top\xcf - c \leq \epsilon
\end{split}
\end{equation}

\subsubsection{Poisson regression}\label{sec:appendix:poissonreg}
Recall that the prediction function $\regression$ is given as
\begin{equation}
\regression(\x) = \exp(\vec{w}^\top\x + b)
\end{equation}
The constraint for exactly obtaining the requested prediction $\ycf$ is
\begin{equation}
\begin{split}
& \regression(\xcf) = \ycf \\
& \Leftrightarrow \exp(\vec{w}^\top\xcf + b) = \ycf \\
& \Leftrightarrow \vec{w}^\top\xcf + b - \log(\ycf) = 0 \\
& \Leftrightarrow \vec{w}^\top\xcf + c = 0
\end{split}
\end{equation}
where
\begin{equation}
c = b - \log(\ycf)
\end{equation}
Finally, we obtain the following set of linear inequality constraints:
\begin{equation}
\begin{split}
& \vec{w}^\top\xcf + c \leq \epsilon \\
& -\vec{w}^\top\xcf - c \leq \epsilon
\end{split}
\end{equation}
where  we introduced the parameter $\epsilon \geq 0$ that specifies the maximum tolerated deviation from the requested prediction - we set $\epsilon = 0$ if we do not allow any deviations.

\subsubsection{Exponential regression}\label{sec:appendix:expreg}
Recall that the prediction function $\regression$ is given as:
\begin{equation}\label{eq:h:expreg}
\regression(\x) = -\frac{1}{\vec{w}^\top\x + b}
\end{equation}
The constraint for a specific prediction $\ycf$ is given as:
\begin{equation}\label{eq:expreg:constraint}
\begin{split}
& \regression(\xcf) = \ycf \\
& \Leftrightarrow -\frac{1}{\vec{w}^\top\xcf + b} = \ycf \\
& \Leftrightarrow \vec{w}^\top\xcf + b + \frac{1}{\ycf} = 0 \\
& \Leftrightarrow \vec{w}^\top\xcf + c = 0
\end{split}
\end{equation}
where
\begin{equation}
c = b + \frac{1}{\ycf}
\end{equation}
Finally, we obtain the following set of linear inequality constraints:
\begin{equation}
\begin{split}
& \vec{w}^\top\xcf + c \leq \epsilon \\
& -\vec{w}^\top\xcf - c \leq \epsilon
\end{split}
\end{equation}
where  we introduced the parameter $\epsilon \geq 0$ that specifies the maximum tolerated deviation from the requested prediction - we set $\epsilon = 0$ if we do not allow any deviations.

\subsection{Gaussian naive bayes}\label{sec:appendix:gnb}
Recall that the prediction function $\classifier$ is given as:
\begin{equation}\label{eq:h:gnb}
\classifier(\x) = \underset{i \,\in\,\set{Y}}{\arg\max}\;\prod_{k=1}^d \N(\x \mid \mu_{ik}, \sigma^2_{ik})\pi_i
\end{equation}
We note that Eq.~\ref{eq:h:gnb} is equivalent to
\begin{equation}\label{eq:h:gnb2}
\classifier(\x) = \underset{i \,\in\,\set{Y}}{\arg\max}\;\sum_{k=1}^d \log\left(\N(\x \mid \mu_{ik}, \sigma^2_{ik})\right) + \log(\pi_i)
\end{equation}
Simplifying the term in Eq.~\ref{eq:h:gnb2} yields
\begin{equation}
\begin{split}
\log(\pi_i) + \sum_{k=1}^d \log\left(\N(\x \mid \mu_{ik}, \sigma^2_{ik})\right) &= \log(\pi_i) + \sum_{k=1}^d \log\left(\frac{1}{\sqrt{2\pi\sigma_{ik}^2}}\right) +\\&\quad\quad \sum_{k=1}^d -\frac{1}{2\sigma_{ik}^2}\pnorm{(\x)_k - \mu_{ik}}_2^2 \\
 &= \log(\pi_i) + \sum_{k=1}^d \log\left(\frac{1}{\sqrt{2\pi\sigma_{ik}^2}}\right) -\\&\quad\quad \sum_{k=1}^d \frac{1}{2\sigma_{ik}^2}\Big((\x)_k^2 + \mu_{ik}^2 - 2(\x)_k\mu_{ik}\Big) \\
 &= c_i - \x^\top\mat{A}_i\x + \vec{q}_i^\top\x
\end{split}
\end{equation}
where
\begin{equation}
c_i = \log(\pi_i) + \sum_{k=1}^d \log\left(\frac{1}{\sqrt{2\pi\sigma_{ik}^2}}\right) - \frac{\mu_{ik}^2}{2\sigma_{ik}^2}
\end{equation}
\begin{equation}
\mat{A}_i = \diag\left(\frac{1}{2\sigma_{ik}^2}\right)
\end{equation}
\begin{equation}
\vec{q}_i = \left(\frac{\mu_{i1}}{\sigma_{i1}^2}, \dots, \frac{\mu_{id}}{\sigma_{id}^2}\right)^\top
\end{equation}
For a sample $\x$, in order to be classified as the $i$-th class, the following set of strict inequalities must hold:
\begin{equation}\label{eq:gnb:inequalities}
\begin{split}
c_i - \x^\top\mat{A}_i\x + \vec{q}_i^\top\x > c_j - \x^\top\mat{A}_j\x + \vec{q}_j^\top\x \quad \forall\,j\neq i
\end{split}
\end{equation}
By rearranging terms in Eq.~\ref{eq:gnb:inequalities}, we get the final constraints
\begin{equation}\label{eq:gnb:finalconstraint}
\x^\top\mat{A}_{ij}\x + \vec{q}_{ij}^\top\x + c_{ij} < 0 \quad \forall\,j\neq i
\end{equation}
where
\begin{equation}
\mat{A}_{ij} = \mat{A}_i - \mat{A}_j = \diag\left(\frac{1}{2\sigma_{ik}^2} - \frac{1}{2\sigma_{jk}^2}\right)
\end{equation}
\begin{equation}
\vec{q}_{ij} = \vec{q}_j - \vec{q}_i = \left(\frac{\mu_{j1}}{\sigma_{j1}^2} - \frac{\mu_{i1}}{\sigma_{i1}^2}, \dots, \frac{\mu_{jd}}{\sigma_{jd}^2} - \frac{\mu_{id}}{\sigma_{id}^2}\right)^\top
\end{equation}
\begin{equation}
\begin{split}
c_{ij} &= c_j - c_i = \log(\pi_j) - \log(\pi_i) +\\&\quad\quad \sum_{k=1}^d \log\left(\frac{1}{\sqrt{2\pi\sigma_{jk}^2}}\right) - \frac{\mu_{jk}^2}{2\sigma_{jk}^2} - \log\left(\frac{1}{\sqrt{2\pi\sigma_{ik}^2}}\right) + \frac{\mu_{ik}^2}{2\sigma_{ik}^2} \\
&= \log\left(\frac{\pi_j}{\pi_i}\right) + \sum_{k=1}^d \log\left(\frac{\sqrt{2\pi\sigma_{ik}^2}}{\sqrt{2\pi\sigma_{jk}^2}}\right) - \frac{\mu_{jk}^2}{2\sigma_{jk}^2} + \frac{\mu_{ik}^2}{2\sigma_{ik}^2}
\end{split}
\end{equation}
Because we can not make any statement about the definiteness of the diagonal matrix $\mat{A}_{ij}$, the constraint Eq.~\ref{eq:gnb:finalconstraint} is a non-convex quadratic inequality constraint.

\subsection{Quadratic discriminant analysis}\label{sec:appendix:qda}
Recall that the prediction function $\classifier$ is given as:
\begin{equation}\label{eq:h:qda}
\classifier(\vec{x}) = \underset{i \,\in\,\set{Y}}{\arg\max}\;\N(\vec{x} \mid \vec{\mu}_i, \mat{\Sigma}_i)\pi_i
\end{equation}
We can rewrite Eq.~\ref{eq:h:qda} as
\begin{equation}
\classifier(\x) = \underset{i \,\in\,\set{Y}}{\arg\max}\;\log\Big(\N(\x \mid \vec{\mu}_i, \mat{\Sigma}_i)\pi_i\Big)
\end{equation}
Working on the $\log$ term yields
\begin{equation}
\begin{split}
\log\Big(\N(\x \mid \vec{\mu}_i, \mat{\Sigma}_i)\pi_i\Big) &= \log\Big(\N(\x \mid \vec{\mu}_i, \mat{\Sigma}_i)\Big) + \log(\pi_i) \\
 &= -\frac{d}{2}\log(2\pi) - \frac{1}{2}\log\big(\det(\mat{\Sigma}_i^{-1})\big) - \frac{1}{2}(\x - \vec{\mu}_i)^\top\mat{\Sigma}_i^{-1}(\x - \vec{\mu}_i)\\&\quad + \log(\pi_i) \\
 &= -\frac{1}{2}\x^\top\mat{\Sigma}_i^{-1}\x + \x^\top\vec{q}_i + c_i
\end{split}
\end{equation}
where
\begin{equation}
\vec{q}_i = \mat{\Sigma}_i^{-1}\vec{\mu}_i
\end{equation}
\begin{equation}
c_i = -\frac{d}{2}\log(2\pi) - \frac{1}{2}\log\big(\det(\mat{\Sigma}_i)\big) -\frac{1}{2}\vec{\mu}_i^\top\mat{\Sigma}_i^{-1}\vec{\mu}_i + \log(\pi_i)
\end{equation}
For a sample $\x$, in order to be classified as the $i$-th class, the following set of strict inequalities must hold:
\begin{equation}\label{eq:qda:constraint}
-\frac{1}{2}\x^\top\mat{\Sigma}_i^{-1}\x + \x^\top\vec{q}_i + c_i > -\frac{1}{2}\x^\top\mat{\Sigma}_j^{-1}\x + \x^\top\vec{q}_j + c_j \quad \forall\,j\neq i
\end{equation}
Rearranging Eq.~\ref{eq:qda:constraint} yields
\begin{equation}\label{eq:qda:finalconstraint}
\frac{1}{2}\x^\top\mat{A}_{ij}\x + \x^\top\vec{q}_{ij} + c_{ij} < 0 \quad \forall\,j\neq i
\end{equation}
where
\begin{equation}
\mat{A}_{ij} = \mat{\Sigma}_i^{-1} - \mat{\Sigma}_j^{-1}
\end{equation}
\begin{equation}
\vec{q}_{ij} = \vec{q}_j - \vec{q}_i = \mat{\Sigma}_j^{-1}\vec{\mu}_j - \mat{\Sigma}_i^{-1}\vec{\mu}_i
\end{equation}
\begin{equation}
\begin{split}
c_{ij} &= c_j - c_i \\
 &= -\frac{d}{2}\log(2\pi) - \frac{1}{2}\log\big(\det(\mat{\Sigma}_j)\big) -\frac{1}{2}\vec{\mu}_j^\top\mat{\Sigma}_j^{-1}\vec{\mu}_j + \log(\pi_j) +\\&\quad \frac{d}{2}\log(2\pi) + \frac{1}{2}\log\big(\det(\mat{\Sigma}_i)\big) + \frac{1}{2}\vec{\mu}_i^\top\mat{\Sigma}_i^{-1}\vec{\mu}_i - \log(\pi_i) \\
  &= \frac{1}{2}\Big(\vec{\mu}_i^\top\mat{\Sigma}_i^{-1}\vec{\mu}_i - \vec{\mu}_j^\top\mat{\Sigma}_j^{-1}\vec{\mu}_j\Big) + \frac{1}{2}\log\left(\frac{\det(\mat{\Sigma}_i)}{\det(\mat{\Sigma}_j)}\right) + \log\left(\frac{\pi_j}{\pi_i}\right)
\end{split}
\end{equation}
The final constraint Eq.~\ref{eq:qda:finalconstraint} is a non-convex quadratic constraint because we can not make any statement about the definiteness of $\mat{A}_{ij}$.

\subsection{Learning vector quantization}\label{sec:appendix:lvq}
Note: The subsequent sections are taken from~\cite{counterfactualslvq}.

\subsubsection{Enforcing a specific prototype as the nearest neighbor}
By using the following set of inequalities, we can force the prototype $\vec{\prototype}_i$ to be the nearest neighbor of the counterfactual $\xcf$ - which would cause $\xcf$ to be classified as $\protolabel_i$: 
\begin{equation}
\dist(\xcf, \vec{\prototype}_i) < \dist(\xcf, \vec{\prototype}_j) \quad \forall\, \vec{\prototype}_j\in\set{P}(\ycf)
\end{equation}
We consider a fixed pair of $i$ and $j$:
\begin{equation}\label{eq:nearestprototypeconstraint}
\begin{split}
& \dist(\xcf, \vec{\prototype}_i) < \dist(\xcf, \vec{\prototype}_j) \\
& \Leftrightarrow \pnorm{\xcf - \vec{\prototype}_i}_{\distmat_i}^2 < \pnorm{\xcf - \vec{\prototype}_j}_{\distmat_j}^2 \\
& \Leftrightarrow (\xcf - \vec{\prototype}_i)^\top{\distmat}_i(\xcf - \vec{\prototype}_i) < (\xcf - \vec{\prototype}_j)^\top{\distmat}_j(\xcf - \vec{\prototype}_j) \\
& \Leftrightarrow \xcf^\top{\distmat}_i\xcf - 2\xcf^\top{\distmat}_i\vec{\prototype}_i + \vec{\prototype}_i^\top{\distmat}_i\vec{\prototype}_i < \xcf^\top{\distmat}_j\xcf - 2\xcf^\top{\distmat}_j\vec{\prototype}_j + \vec{\prototype}_j^\top{\distmat}_i\vec{\prototype}_j \\
& \Leftrightarrow \xcf^\top{\distmat}_i\xcf - \xcf^\top{\distmat}_j\xcf - 2\xcf^\top{\distmat}_i\vec{\prototype}_i + 2\xcf^\top{\distmat}_j\vec{\prototype}_j + \vec{\prototype}_i^\top{\distmat}_i\vec{\prototype}_i - \vec{\prototype}_j^\top{\distmat}_i\vec{\prototype}_j < 0 \\
& \Leftrightarrow \xcf^\top({\distmat}_i - {\distmat}_j)\xcf + \xcf^\top(-2{\distmat}_i\vec{\prototype}_i + 2{\distmat}_j\vec{\prototype}_j) + (\vec{\prototype}_i^\top{\distmat}_i\vec{\prototype}_i - \vec{\prototype}_j^\top{\distmat}_i\vec{\prototype}_j) \\
& \Leftrightarrow \frac{1}{2}\xcf^\top({\distmat}_i - {\distmat}_j)\xcf + \frac{1}{2}\xcf^\top({\distmat}_j\vec{\prototype}_j - {\distmat}_i\vec{\prototype}_i) + \frac{1}{2}(\vec{\prototype}_i^\top{\distmat}_i\vec{\prototype}_i - \vec{\prototype}_j^\top{\distmat}_i\vec{\prototype}_j) < 0 \\
& \Leftrightarrow \frac{1}{2}\xcf^\top\mat{A}_{ij}\xcf + \xcf^\top\vec{q}_{ij} + c_{ij} < 0
\end{split}
\end{equation}
where
\begin{equation}
\mat{A}_{ij} = {\distmat}_{i} - {\distmat}_{j}
\end{equation}
\begin{equation}
\vec{q}_{ij} = \frac{1}{2}\left({\distmat}_j\vec{\prototype}_{j} - {\distmat}_i\vec{\prototype}_{i}\right)
\end{equation}
\begin{equation}
c_{ij} = \frac{1}{2}\left(\vec{\prototype}_{i}\top{\distmat}_{i}\vec{\prototype}_{i} - \vec{\prototype}_{j}\top{\distmat}_{j}\vec{\prototype}_{j}\right)
\end{equation}
\mbox{}\\
If we only have one global distance matrix $\distmat$, we find that $\mat{A}_{ij}=\mat{0}$ and the inequality Eq.~\ref{eq:nearestprototypeconstraint} simplifies:
\begin{equation}
\begin{split}
& \dist(\x, \vec{\prototype}_i) < \dist(\x, \vec{\prototype}_j) \\
& \Leftrightarrow \xcf^\top\vec{q}_{ij} + c_{ij} < 0
\end{split}
\end{equation}
where
\begin{equation}
\vec{q}_{ij} = \frac{1}{2}{\distmat}\left(\vec{\prototype}_{j} - \vec{\prototype}_{i}\right)
\end{equation}
\begin{equation}
c_{ij} = \frac{1}{2}\left(\vec{\prototype}_{i}^\top\distmat\vec{\prototype}_{i} - \vec{\prototype}_{j}^\top\distmat\vec{\prototype}_{j}\right)
\end{equation}
\mbox{}\\
If we do not use a custom distance matrix, we have $\distmat=\I$ and Eq.~\ref{eq:nearestprototypeconstraint} becomes:
\begin{equation}
\begin{split}
& \dist(\x, \vec{\prototype}_i) < \dist(\x, \vec{\prototype}_j) \\
& \Leftrightarrow \xcf^\top\vec{q}_{ij} + c_{ij} < 0
\end{split}
\end{equation}
where
\begin{equation}
\vec{q}_{ij} = \frac{1}{2}\left(\vec{\prototype}_{j} - \vec{\prototype}_{i}\right)
\end{equation}
\begin{equation}
c_{ij} = \frac{1}{2}\left(\vec{\prototype}_{i}^\top\vec{\prototype}_{i} - \vec{\prototype}_{j}^\top\vec{\prototype}_{j}\right)
\end{equation}

\subsection{Minimizing the Euclidean distance}\label{sec:appendix:mineuclideandistance}
Minimizing the Euclidean distance (Eq.~\ref{eq:lvq_generaldist}) yields a \textit{quadratic} objective.

First, we expand the Euclidean distance (Eq.~\ref{eq:lvq_generaldist}):
\begin{equation}
\begin{split}
\pnorm{\xcf - \x}_2^2 &= (\xcf - \x)^\top(\xcf - \x) \\
&= \xcf^\top\xcf - \xcf^\top\x - \x^\top\xcf + \x^\top\x \\
&= \xcf^\top\xcf - 2\x^\top\xcf + \x^\top\x \\
\end{split}
\end{equation}
Next, we note that that we can drop the constant $\x^\top\x$ when optimizing with respect to $\xcf$:
\begin{equation}
\begin{split}
& \underset{\xcf \,\in\, \RN^d}{\min}\; \pnorm{\xcf - \x}_2^2 \\
& \Leftrightarrow \\
& \underset{\xcf \,\in\, \RN^d}{\min}\; \frac{1}{2}\xcf^\top\xcf - \x^\top\xcf
\end{split}
\end{equation}

\subsection{Minimizing the weighted Manhattan distance}\label{sec:appendix:minmanhattandistance}
Minimizing the weighted Manhattan distance (Eq.~\ref{eq:weighted_l1}) yields a \textit{linear} objective.

First, we transform the problem of minimizing the weighted Manhattan distance (Eq.~\ref{eq:weighted_l1}) into epigraph form:
\begin{equation}
\begin{split}
& \underset{\xcf \,\in\, \RN^d}{\min}\; \sum_j \alpha_j \cdot |(\xcf)_j - (\x)_j| \\
& \Leftrightarrow \\
& \underset{\xcf \,\in\, \RN^d, \beta\,\in\, \RN}{\min}\; \beta \\
& \quad \text{s.t.} \quad \sum_j \alpha_j \cdot |(\xcf)_j - (\x)_j| \leq \beta \\
& \quad \quad \quad \beta \geq 0
\end{split}
\end{equation}
Next, we separate the dimensions:
\begin{equation}
\begin{split}
& \underset{\xcf \,\in\, \RN^d, \beta\,\in\, \RN}{\min}\; \beta \\
& \quad \text{s.t.} \quad \sum_j \alpha_j \cdot |(\xcf)_j - (\x)_j| \leq \beta \\
& \quad \quad \quad \beta \geq 0 \\
& \Leftrightarrow \\
& \underset{\xcf, \vec{\beta} \,\in\, \RN^d}{\min}\; \sum_j (\vec{\beta})_j \\
& \quad \text{s.t.} \quad \alpha_j \cdot |(\xcf)_j - (\x)_j| \leq (\vec{\beta})_j \quad \forall\,j \\
& \quad \quad \quad (\vec{\beta})_j \geq 0 \quad \forall\,j
\end{split}
\end{equation}
After that, we remove the absolute value function:
\begin{equation}
\begin{split}
& \underset{\xcf, \vec{\beta} \,\in\, \RN^d}{\min}\; \sum_j (\vec{\beta})_j \\
& \quad \text{s.t.} \quad \alpha_j \cdot |(\xcf)_j - (\x)_j| \leq (\vec{\beta})_j \quad \forall\,j \\
& \quad \quad \quad (\vec{\beta})_j \geq 0 \quad \forall\,j \\
& \Leftrightarrow \\
& \underset{\xcf, \vec{\beta} \,\in\, \RN^d}{\min}\; \sum_j (\vec{\beta})_j \\
& \quad \text{s.t.} \quad \alpha_j (\xcf)_j - \alpha_j (\x)_j \leq (\vec{\beta})_j \quad \forall\,j \\
& \quad \quad \quad -\alpha_j (\xcf)_j + \alpha_j (\x)_j \leq (\vec{\beta})_j \quad \forall\,j \\
& \quad \quad \quad (\vec{\beta})_j \geq 0 \quad \forall\,j
\end{split}
\end{equation}
Finally, we rewrite everything in matrix-vector notation:
\begin{equation}
\begin{split}
& \underset{\xcf, \vec{\beta} \,\in\, \RN^d}{\min}\;\vec{1}^\top\vec{\beta} \\
& \text{s.t.} \\
& \mat{\Upsilon}\xcf - \mat{\Upsilon}\x \leq \vec{\beta} \\
& -\mat{\Upsilon}\xcf + \mat{\Upsilon}\x \leq \vec{\beta} \\
& \vec{\beta} \geq \vec{0}
\end{split}
\end{equation}
where
\begin{equation}
\mat{\Upsilon} = \diag(\alpha_j)
\end{equation}

\subsection{Solving a non-convex QCQP}
Solving a non-convex QCQP is known to be NP-hard~\cite{Boyd2004,park2017general}.

In section~\ref{sec:appendix:nonconvexqcqp} we discuss a method for approximately solving a non-convex QCQP and in section~\ref{sec:appendix:qcqp1constraint} we describe how to solve the special case of a non-convex QCQP having a single constraint.

\subsubsection{Approximately solving a non-convex QCQP}\label{sec:appendix:nonconvexqcqp}
Recall the non-convex quadratic constraint:
\begin{equation}\label{eq:nonconvexquadratic}
\frac{1}{2}\xcf^\top\mat{A}_{ij}\xcf + \xcf^\top\vec{q}_{ij} + r_{ij} \leq 0
\end{equation}
In this paper, we always defined the matrix $\mat{A}_{ij}$ as the difference of two s.p.s.d. matrices $\mat{A}_{i}$ and $\mat{A}_{j}$:
\begin{equation}\label{eq:matrixdiff}
\mat{A}_{ij} = \mat{A}_{i} - \mat{A}_{j}
\end{equation}
By making use of Eq.~\ref{eq:matrixdiff}, we can rewrite Eq.~\ref{eq:nonconvexquadratic} as:
\begin{equation}
\begin{split}
& \frac{1}{2}\xcf^\top\mat{A}_{i}\xcf + \xcf^\top\vec{q}_{ij} + r_{ij} - \frac{1}{2}\xcf^\top\mat{A}_{j}\xcf \leq 0 \\
& \Leftrightarrow f(\xcf) - g(\xcf) \leq 0
\end{split}
\end{equation}
where
\begin{equation}
f(\xcf) = \frac{1}{2}\xcf^\top\mat{A}_{i}\xcf + \xcf^\top\vec{q}_{ij} + r_{ij} 
\end{equation}
\begin{equation}
g(\xcf) = \frac{1}{2}\xcf^\top\mat{A}_{j}\xcf
\end{equation}
Under the assumption that our regularization function $\regularization()$ is a convex function\footnote{The weighted Manhattan distance and the Euclidean distance are convex functions!}, we can rewrite a generic version of the non-convex QCQP Eq.~\ref{eq:cf:lgmlvq} as follows:
\begin{equation}\label{eq:dcp}
\begin{split}
& \underset{\xcf \,\in\, \RN^d}{\min}\;\regularization(\xcf, \x) \\
& \text{s.t.} \\
& f(\xcf) - g(\xcf) \leq 0
\end{split}
\end{equation}
Because $\mat{A}_{i}$ and $\mat{A}_{j}$ are s.p.s.d. matrices, we know that $f(\xcf)$ and $g(\xcf)$ are convex functions. Therefore, Eq.~\ref{eq:dcp} is a difference-of-convex program (DCP).

This allows us to use the penalty convex-concave procedure (CCP)~\cite{park2017general} for computing an approximate solution of Eq.~\ref{eq:dcp}, yielding an approximate solution of the original non-convex QCQP. For using the penalty CCP, we need the first order Taylor approximation of $g(\xcf)$ around a current point $\x_k$:
\begin{equation}
\begin{split}
\hat{g}(\xcf)_{\x_k} &= g(\x_k) + (\nabla_{\xcf}g)(\x_k)^\top(\xcf - \x_k) \\
&= \frac{1}{2}\x_k^\top\mat{A}_{j}\x_k + (\mat{A}_{j}\x_k)^\top(\xcf - \x_k) \\
&= (\mat{A}_{j}\x_k)^\top\xcf + \frac{1}{2}\x_k^\top\mat{A}_{j}\x_k -(\mat{A}_{j}\x_k)^\top\x_k \\
&= \vec{\rho}_{jk}^\top\xcf + \tilde{c}_{jk}
\end{split}
\end{equation}
where
\begin{equation}
\vec{\rho}_{jk} = \mat{A}_{j}\x_k
\end{equation}
\begin{equation}
\tilde{c}_{jk} = -\frac{1}{2}\x_k^\top\mat{A}_{j}\x_k
\end{equation}
In order to run the convex-concave procedure, we have to provide an initial (feasible) solution. We could either use the original data point as an initial \textit{infeasible} solution, some data point yielding the requested prediction as an initial \textit{feasible} solution or some other ``smart'' initialization.

As an alternative, we could use other methods for approximately computing a solution of the non-convex QCQP like the Suggest-Improve framework~\cite{park2017general} - actually, the methods we described in the previous paragraph is an instance of the Suggest-Improve framework.

\subsubsection{Solving a non-convex QCQP with just one constraint}\label{sec:appendix:qcqp1constraint}
We consider the general QCQP
\begin{equation}\label{eq:qcqp1constraint}
\begin{split}
&\underset{\xcf \,\in\, \RN^d}\min\;\xcf^\top\mat{Q}\xcf + \vec{q}^\top\xcf + c \\
& \text{s.t.} \\
& \xcf^\top\mat{A}\xcf + \vec{b}^\top\xcf + r \leq 0
\end{split}
\end{equation}
where $\mat{Q}, \mat{A} \in \RN^{d \times d}$, $\vec{q}, \vec{b} \in \RN^d$ and $c, r \in \RN$.

If $\mat{Q}$ and $\mat{A}$ are not symmetric positive semi-definite, Eq.~\ref{eq:qcqp1constraint} is a non-convex QCQP. However, despite the non-convexity, we can solve Eq.~\ref{eq:qcqp1constraint} efficiently by solving the dual of Eq.~\ref{eq:qcqp1constraint} and observing that the duality gap is zero~\cite{Boyd2004} - under the assumption that Eq.~\ref{eq:qcqp1constraint} is strictly feasible\footnote{We can always achieve strict feasibility by moving a non-strict feasible point away from the decision boundary.}.

Therefore, solving Eq.~\ref{eq:qcqp1constraint} is equivalent to solving the following semi-definite program (SDP)~\cite{Boyd2004}:
\begin{equation}
\begin{split}
& \underset{\mat{X} \,\in\,\set{S}^d, \xcf\,\in\,\RN^d}{\arg\min}\;\trace(\mat{Q}\mat{X}) + \vec{q}^\top\xcf + c \\
& \text{s.t.}\\
& \trace\left(\mat{A}\mat{X}\right) + \vec{b}^\top\xcf + r \leq 0 \\
& \begin{pmatrix} \mat{X} & \xcf \\ \xcf^\top & 1 \end{pmatrix} \succeq 0
\end{split}
\end{equation}
where we introduced an additional variable\footnote{$\set{S}^d$ denotes the set of symmetric $\RN^{d \times d}$ matrices} $\mat{X}$ that can be discarded afterwards.


\begin{footnotesize}




\bibliographystyle{unsrt}
\bibliography{bibliography}

\end{footnotesize}


\end{document}